\documentclass[journal,twoside,web]{ieeecolor}

\usepackage[dvipsnames]{xcolor}
\usepackage{generic}
\usepackage{cite}
\usepackage{amsmath,amssymb,amsfonts}
\usepackage{graphicx}
\usepackage{algorithmic}
\usepackage{algorithm,algorithmic}

\usepackage{hyperref}
\hypersetup{hidelinks}

\usepackage{textcomp}
\usepackage{forest}
\usepackage[normalem]{ulem}
\useunder{\uline}{\ul}{}

\usepackage{tabularx,longtable}
\usepackage{multirow}
\usepackage{booktabs}
\usepackage{makecell}
\usepackage{amssymb}
\usepackage{pifont}

\newcommand{\xmark}{\ding{55}}

\let\labelindent\relax
\usepackage{enumitem}

\def\BibTeX{{\rm B\kern-.05em{\sc i\kern-.025em b}\kern-.08em
    T\kern-.1667em\lower.7ex\hbox{E}\kern-.125emX}}
\begin{document}
\title{Anatomy-Guided Radiology Report Generation with Pathology-Aware Regional Prompts}
\author{Yijian Gao, Dominic Marshall, Xiaodan Xing, Junzhi Ning, Giorgos Papanastasiou, Guang Yang*, \IEEEmembership{Senior Member, IEEE} and Matthieu Komorowski*
\thanks{This study was supported in part by the ERC IMI (101005122), the H2020 (952172), the MRC (MC/PC/21013), the Royal Society (IEC$\backslash$NSFC$\backslash$211235), the NVIDIA Academic Hardware Grant Program, the SABER project supported by Boehringer Ingelheim Ltd, NIHR Imperial Biomedical Research Centre (RDA01), Wellcome Leap Dynamic Resilience, UKRI guarantee funding for Horizon Europe MSCA Postdoctoral Fellowships (EP/Z002206/1), and the UKRI Future Leaders Fellowship (MR/V023799/1). Dominic Marshall is supported by an MRC clinical research training fellowship (award MR/Y000404/1) and the Mittal Fund at Cleveland Clinic Philanthropy. (\textit{Corresponding Author: Guang Yang.})}
\thanks{Yijian Gao and Junzhi Ning were with the Department of Computing, Imperial College London, London SW7 2AZ, UK (e-mail: y.gao23@imperial.ac.uk, j.ning23@imperial.ac.uk). }
\thanks{Xiaodan Xing and Guang Yang are with the Bioengineering Department and Imperial-X, Imperial College London, London W12 7SL, UK (e-mail: x.xing@imperial.ac.uk, g.yang@imperial.ac.uk).}
\thanks{Giorgos Papanastasiou is with the Archimedes Unit, Athena Research Centre, 15125 Athens, Greece (e-mail: g.papanastasiou@athenarc.gr).}
\thanks{Dominic Marshall and Matthieu Komorowski are with the Department of Surgery and Cancer, Imperial College London,  London SW7 2AZ, UK (e-mail: dominic.marshall12@imperial.ac.uk, m.komorowski14@imperial.ac.uk).}
\thanks{\textit{* Co-last senior authors.}}
\vspace{-25pt}
}
\maketitle

\begin{abstract}

Radiology reporting generative AI holds significant potential to alleviate clinical workloads and streamline medical care. However, achieving high clinical accuracy is challenging, as radiological images often feature subtle lesions and intricate structures. Existing systems often fall short, largely due to their reliance on fixed size, patch-level image features and insufficient incorporation of pathological information. This can result in the neglect of such subtle patterns and inconsistent descriptions of crucial pathologies. To address these challenges, we propose an innovative approach that leverages pathology-aware regional prompts to explicitly integrate anatomical and pathological information of various scales, significantly enhancing the precision and clinical relevance of generated reports. We develop an anatomical region detector that extracts features from distinct anatomical areas, coupled with a novel multi-label lesion detector that identifies global pathologies. Our approach emulates the diagnostic process of radiologists, producing clinically accurate reports with comprehensive diagnostic capabilities. Experimental results show that our model outperforms previous state-of-the-art methods on most natural language generation and clinical efficacy metrics, with formal expert evaluations affirming its potential to enhance radiology practice.

\end{abstract}

\begin{IEEEkeywords}
 Radiology Report Generation, Prompt Learning, Anatomy Detection, Lesion localization
\end{IEEEkeywords}

\section{Introduction}\label{sec:introduction}

\IEEEPARstart{E}{very} year, the demand for medical imaging, its interpretation and reporting exceeds the pace of growth in the workforce. Chest radiographs are the most common form of imaging and serve as the initial investigation for a vast array of medical presentations, with around 2 billion exams conducted globally each year \cite{akhter2023ai}. This escalating workload exceeds available resources, causing reporting delays that can result in clinicians making critical decisions without specialist input \cite{ukchestx-ray}. Therefore, a well-optimized automated reporting tool holds immense potential to streamline clinical workflows and improve the efficiency and standardization of medical care.

Drawing inspiration from the success of generative AI in natural image captioning \cite{xu2015show, lu2017knowing}, numerous studies have adopted the popular encoder-decoder architecture, involving a visual encoder like the Convolutional Neural Network (CNN) \cite{krizhevsky2012imagenet} to extract image features, and a language decoder like the Transformer \cite{vaswani2017attention} to generate a free-text report, incorporating various assistive techniques to enhance the report quality \cite{survey}.

Nevertheless, the performance of existing radiology report generation methods fall short of real-world diagnostic applications, particularly in terms of clinical accuracy \cite{survey, he2024survery}. Aligning comprehensive descriptions with precise diagnostic outcomes remains a pivotal challenge, mainly due to the intrinsic characteristics of radiological images (e.g., subtle lesions) and commonly lengthy reports. 

Existing approaches \cite{jin2023promptmrg, CAMANet, r2gen} typically rely on patch-level feature extraction with fixed grids in encoders, neglecting critical anatomical nuances and hampering the integration of holistic diagnostic views. This limitation frequently leads to incomplete and inconsistent descriptions, including redundant narratives of the same area, errors in laterality, or insufficient differentiation between organs.

Traditional language decoders, which primarily process only extracted visual features, often fail to fully harness the diagnostic information of radiological images. Additionally, the potential of prompt guidance to enhance the generalization capabilities of language models remains under-explored \cite{Liu2023promptsurvey}. By incorporating pathological information as prompt guidance for the language decoder \cite{jin2023promptmrg, wangPrompt}, the model is equipped with high-level semantics that describe specific diagnostic details, enhancing its clinical relevance and accuracy.

To bridge the gap between anatomical and pathological information, we introduce a novel approach inspired by radiologists' diagnostic practices: pathology-aware regional prompts. This method effectively integrates both types of information, mimicking how radiologists correlate overall pathology with specific anatomical regions. Unlike traditional methods that primarily employ patch-level feature extraction in the encoder, our approach develops an anatomical region detector to extract anatomy-level visual features from 29 distinct regions under various scales, accommodating the significant variations in anatomical structure scales in chest X-ray (CXR) images.

Moreover, we introduce a novel multi-label lesion detector into the radiology report generation process, capable of identifying multiple pathologies within a single bounding box. Detected lesions are assigned to overlapping anatomical regions, generating pathology-aware regional prompts that explicitly guide the report decoder with fine-grained diagnostic results. This integration of anatomy-level visual features and prompt guidance significantly enhances the precision and clinical utility of the generated reports. In summary, the key contributions of this work include:

\begin{enumerate}[leftmargin=*,align=left]
\item We propose to mimic the working pattern of radiologists by jointly integrating anatomical and pathological information of various scales via pathology-aware regional prompts, gaining advanced clinically precise reports.

\item We develop an anatomical region detector to serve as the visual encoder, identifying and extracting anatomy-level visual features for 29 distinct regions within a CXR image.

\item We introduce a multi-label lesion detector to enhance the report generation decoder, which is capable of detecting multiple pathologies within a single bounding box.

\item Extensive comparisons on the MIMIC-CXR-JPG with Chest ImaGenome datasets demonstrate the superiority of our model over previous state-of-the-art report generation approaches, with its clinical accuracy and effectiveness further confirmed by formal expert evaluations.
\end{enumerate}

\section{Related Work}
\subsection{Encoder-Decoder Architecture}
The conventional encoder-decoder workflow is widely adopted in existing radiology report generation methods. This architecture integrates a visual encoder for image feature extraction and a language decoder for generating diagnostic reports, supplemented by innovative assistive techniques to optimize the process. 

\subsubsection{Visual Encoders}
Convolutional Neural Networks (CNNs) \cite{krizhevsky2012imagenet} are widely used as feature extractors in computer vision, showing effectiveness in various medical tasks \cite{santosh2022advances}. Wang et al. \cite{CAMANet} used DenseNet121 \cite{huang2017densely} to derive initial patch visual tokens, supporting their class activation map-guided network. ResNet models \cite{ResNet50} are also widely adopted by many studies \cite{yi_tsget_2024, divya_memory_2024, peng_eye_2024}. Recent work has integrated the pre-trained Swin Transformer \cite{liu2021swin} as a visual encoder in models described by \cite{wangPrompt, CGFTrans, pan_s3-net_2024}, whereas the Vision Transformer (ViT) \cite{visiontransformer} has been employed similarly in \cite{jin2023promptmrg, XrayGPT, METransformer}. The Swin Transformer excels at capturing fine-grained, scale-invariant features, whereas the ViT provides a broader understanding of the global context. These attributes enhance diagnostic precision and efficiency in clinical settings.

However, conventional encoders, which rely on fixed-size, patch-level feature extraction, may inadequately represent critical imaging anomalies in radiology. This limitation arises because, unlike natural images, medical images often require focused analysis of specific regions to mirror the diagnostic process of radiologists.

\subsubsection{Language Decoders}
Traditionally, the task of decoding visual features and generating free-text reports was approached using Recurrent Neural Networks (RNNs), such as Long Short-Term Memory (LSTM) \cite{hochreiter1997long} and Gated Recurrent Unit (GRU) \cite{GRU} networks. For instance, Liu et al. \cite{survey92} and Zhang et al. \cite{MIRQI} utilized an LSTM-based sentence decoder to output topic vectors and stop signals, thereby guiding the subsequent word decoder to generate the report word-by-word.

Subsequently, the adoption of Transformer-based models as language decoders including BERT \cite{BERT}, GPT-2 Medium \cite{GPT-2-medium}, and the vanilla Transformer \cite{vaswani2017attention} has become increasingly prevalent \cite{r2gen, r2gencmn, METransformer, clinical-bert}. This shift from RNNs to Transformers facilitates the decoupling of temporal dependencies, allowing for the parallel processing of sequential data and mitigating the issue of vanishing gradients \cite{survey103}. However, challenges such as generating lengthy and complex medical details, along with biases in visual and textual data, critically impact the performance in generating radiology reports.

In recent years, radiology report generation has been transformed by Vision-Language Models (VLMs), which integrate pre-trained image encoders with large language models (LLMs) fine-tuned for radiological terminology. This approach often includes a modality transfer component that effectively bridges the gap between visual and linguistic information.

Notably, CheXagent \cite{chexagent}, a foundation model with 8B parameters, combines the Mistral-7B-v0.1 \cite{jiang2023mistral7b} with the EVA-CLIP-g image encoder \cite{evaclip}. Extensively refined on 28 public datasets, It has demonstrated exceptional CXR analysis abilities. Similar advancements in models like MedDr \cite{meddr}, Liu et al.\cite{liu2024bootstrapping}, and XrayGPT \cite{XrayGPT} also underscore the significant capabilities in image-text alignment and accurate report generation. However, in our experiments, models based on LLMs 
underperformed compared to non-LLM models, primarily due to a lack of adequate high-quality image-text pairs and potential hallucinations.

\subsection{Anatomical Region Guided Report Generation}

An anatomical region refers to a body area defined by specific landmarks and structures, such as the chest, abdomen, or more specific areas like the lung fields or heart. Typically, radiology reports consist of multiple sentences, each describing observations from different regions. 

To better reflect radiologists' workflow, many studies \cite{tanida2023interactive, Serrafinding, scenegraph-wang} have transitioned from patch-level to region-level encoders, extracting features from anatomical regions. This shift allows for the recognition of fine-grained morphological features at various scales, significantly improving the clinical relevance of the visual guidance provided to language decoders.

Specifically, Tanida et al.\cite{tanida2023interactive} developed a region-guided report generation (RGRG) model that employs Faster R-CNN \cite{frcnn} to initially detect anatomical regions. This model generates short, consistent sentences for each region, thus enhancing the report's completeness and consistency. It uses a modified GPT-2 Medium \cite{GPT-2-medium} language decoder, surpassing previous state-of-the-art (SOTA) methods on various metrics and enhancing the explainability and interactivity of the report generation process. However, RGRG employs a region selector during inference to decide if a detected region should be described, potentially introducing bottlenecks if the classifier underperforms. Additionally, by generating sentences for each region separately, it tends to produce contradictory descriptions that may mislead subsequent treatments.

Serra et al. \cite{Serrafinding} proposed a refined Faster R-CNN model that integrates finding detection with anatomical localization. By extracting finding-aware anatomical tokens, their method enables the extraction of triples that link anatomical structures to specific findings, serving as input to the language model. This approach significantly enhances the accuracy, completeness, and clinical explainability of radiology reports. However, anatomy-guided methods typically lack high-level medical knowledge or pathological information during decoding, highlighting potential areas for improvement.

\subsection{Prompt Guided Report Generation}
In the realm of generative AI and natural language processing, prompt learning has significantly enhanced the generalization ability of language models \cite{Liu2023promptsurvey}. However, its application in radiology report generation remains limited, as most language decoders only process extracted visual features.

Jin et al. \cite{jin2023promptmrg} initiated an exploration using diagnostic results from a disease classification branch to guide report generation with diagnostic-driven token prompts. They improved classification performance by employing cross-modal feature enhancement, retrieving similar reports from a database to diagnose query images. Utilizing a BERT-base \cite{BERT} language decoder, their approach excelled in clinical efficacy (CE) metrics and was competitive in natural language generation (NLG) metrics. They also tested various prompt types including text, average pooled features, and embedded prompts, underscoring the effectiveness of token prompts in report generation.

Additionally, Wang et al. introduced a similar PromptRRG approach \cite{wangPrompt}, employing disease-enriched prompts generated automatically from fixed templates and classification results. The proposed prompt effectively distilled essential medical knowledge for report generation using a pre-trained Roberta model \cite{liu2019roBERTa}, achieving outstanding performance in both NLG and CE metrics without adding trainable parameters.

Related works have shown the effectiveness of using well-designed medical prompts to guide the language decoder with domain-specific knowledge \cite{qin2022medicalprompy}. However, most methods primarily depend on disease classifiers with limited classes, neglecting the crucial aspect of pathology localization in radiological practice. As an enhancement to previous methods, Jin et al. \cite{jin2023promptmrg} suggest that enriching prompt guidance with more comprehensive disease information could significantly improve the completeness and accuracy of the reports.

\begin{figure*}[htbp]
\centerline{\includegraphics[draft=false, width=\textwidth]{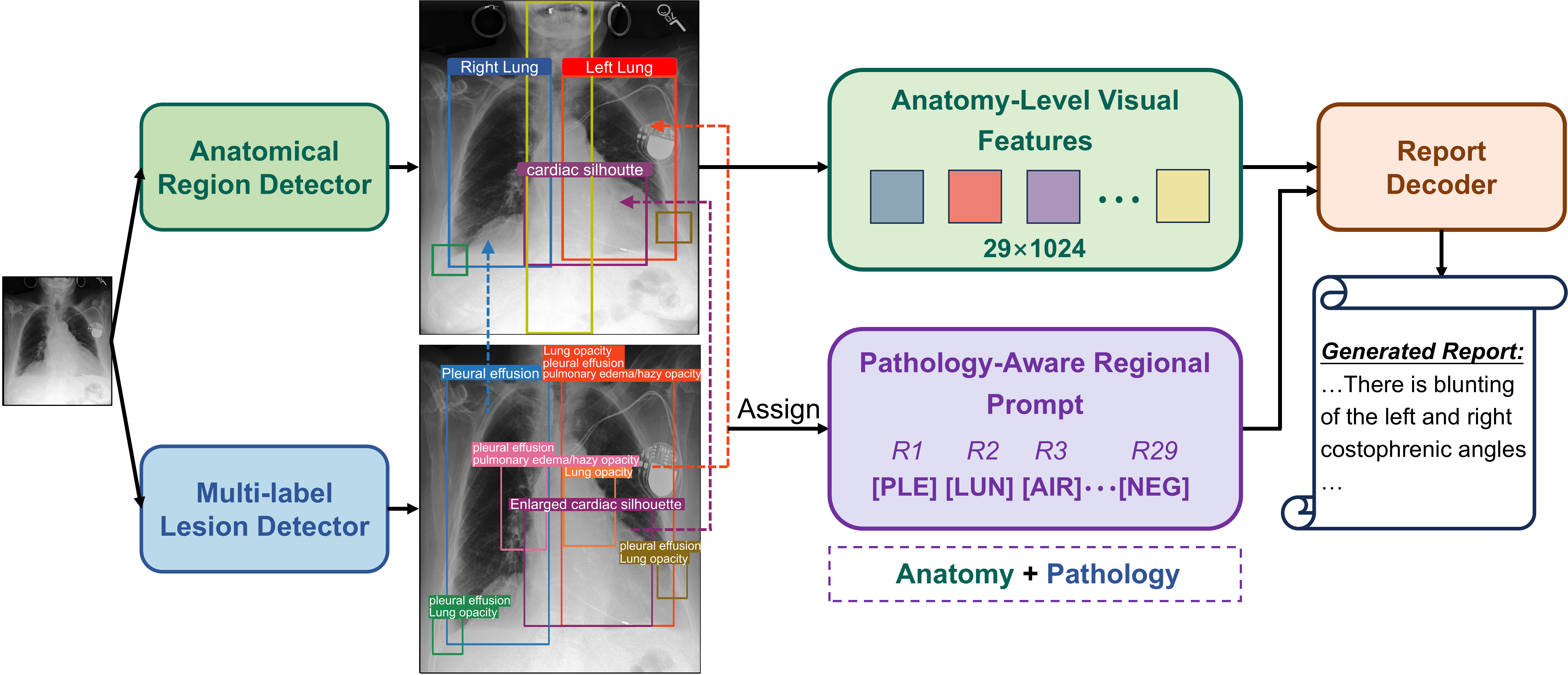}}
\caption{Pipeline of the proposed system. Initially, the anatomical region detector identifies and extracts visual features from 29 regions, with the first six shown in the figure. Simultaneously, the multi-label lesion detector identifies global pathologies, assigning multiple lesions to a single bbox. Pathology-aware regional prompts are generated by mapping lesion bboxes to corresponding anatomical regions based on overlap, each tagged with a lesion token. Finally, the report decoder is explicitly guided by both the anatomy-level visual features and the prompt guidance to generate clinically coherent and accurate radiology reports.} 
\label{fig1-pipeline}
\end{figure*}

\section{Methods}
\subsection{Pipeline Overview}

In radiology report generation, our method mirrors the diagnostic process of radiologists. A radiologist will systematically review all anatomical areas of the chest X-ray (CXR) image, noting each finding and its specific location. This data is then synthesized into an overall impression, listing relevant positive and negative findings by region. Typically, this will be summarized with a summary and sometimes a suggested differential diagnosis.

Our model, depicted in Fig. \ref{fig1-pipeline}, explicitly integrates anatomical and pathological information across diverse scales. It starts with an anatomical region detector that identifies 29 distinct regions and extracts anatomy-level visual features. In contrast to previous methods that rely on fixed-size patch-level features, our approach utilizes anatomy-specific, region-based image features, substantially improving the examination of anatomical observations.

Simultaneously, a multi-label lesion detector is employed to identify pathologies on a global scale. Unlike traditional object detectors, which typically associate each bounding box (bbox) with a single class, our detector can recognize multiple findings within a single bbox, more accurately reflecting radiologists' diagnostic methods.

The outputs from both detectors are then associated based on their spatial locations to generate pathology-aware regional prompts, providing region-specific pathological information. Finally, the report generator employs the anatomy-level visual features alongside the prompt guidance to produce medically relevant descriptions of the CXR image.

The pipeline is trained in two stages for end-to-end inference. The anatomical region detector and multi-label lesion detector are initially trained separately, and the entire model is then trained end-to-end to generate reports with both detectors frozen in the second stage.

\subsection{Anatomical Region Detector}

This component functions as the visual encoder, aimed at accurately detecting 29 distinct anatomical regions in a chest X-ray. We utilize Faster R-CNN \cite{frcnn} with a ResNet-50 \cite{ResNet50} backbone, pre-trained on ImageNet \cite{krizhevsky2012imagenet}, to extract image features. These features are processed by a Region Proposal Network (RPN) to generate object proposals, represented as bounding boxes (bbox):
\begin{equation} r_k = (r_{k,x1}, r_{k,y1}, r_{k,x2}, r_{k,y2}), k=1,2,\ldots,29, \end{equation} 
where $k$ indexes each proposal, and $(x_i, y_i)$ are the top-left and bottom-right bbox coordinates.

Following \cite{tanida2023interactive}, we identify 'top' region proposals and employ a Region of Interest (RoI) pooling layer to produce uniformly sized feature maps, noted as $V_k\in\mathbb{R}^{2048 \times H \times W}$. Subsequently, a 2D average pooling layer with linear transformation is used to reduce the spatial dimensions and derive anatomy-level feature vectors  $V_{k} \in \mathbb{R}^{29 \times 1024}$.

\subsection{Multi-Label Lesion Detector} \label{Lesion_Detector}
\subsubsection{Label Squeeze}

This module enhances the encoder-decoder process by integrating a lesion detector that identifies global pathologies from CXR images. Unlike traditional object detection tasks with single labels per bbox, CXRs often display multiple lesions at the same location. To address this complexity, we modify the YOLOv5x model \cite{ultralytics2021yolov5} to facilitate multi-label training, utilizing its native capabilities and introducing a label squeeze function in the loss calculation process.

Originally, the label vector $l$ for an image is shaped as $[N, 5]$, representing $N$ bboxes, each labeled as $\{class, bbox_i\}$ with $bbox_i = \{x, y, w, h\}$. In our multi-label approach, each bbox may contain several classes. Therefore, we define a class vector $c_i$ of length $C$, where $C$ is the total number of possible classes. The vector is populated according to the class labels associated with $bbox_i$. Consequently, the modified label vector $l'$ reshapes to $[M, C+4]$, with each entry $l'_i = \{c_i, bbox_i\}$, where $M$ is the count of unique bboxes per image. 

Therefore, the bbox classification loss for $bbox_i$ is given by:
\begin{equation}
    L_{\text{cls}} = \sum_{i=1}^M \sum_{j=1}^C -\left( c_{i}[j] \cdot \log(\hat{c}_{i}[j])+(1 - c_{i}[j]) \cdot \log(1 - \hat{c}_{i}[j]) \right),
\end{equation}
where $\hat{c}$ is the predicted class vector.

The objectness loss $L_{\text{obj}}$ measures the model's confidence in detecting any object within a bounding box using Binary Cross-Entropy (BCE) loss between the predicted object presence score and the actual presence (1 for object presence and 0 otherwise). The bbox regression loss $L_{\text{box}}$ calculates the accuracy of the predicted bbox coordinates compared to the ground truth using Mean Squared Error (MSE) loss for $bbox_i$. Finally, the total loss for the multi-label lesion detector combines these elements:

\begin{equation}
    L = \lambda_{\text{cls}} L_{\text{cls}} + \lambda_{\text{obj}} L_{\text{obj}} + \lambda_{\text{box}} L_{\text{box}},
\end{equation}
with $\lambda_{\text{cls}}, \lambda_{\text{obj}}, \lambda_{\text{box}}$ as the respective weights for each loss component.

\subsubsection{Class Reduction}\label{class-reduction}

\begin{figure}[htbp]
  \centering
  \includegraphics[draft=false, width=\columnwidth]{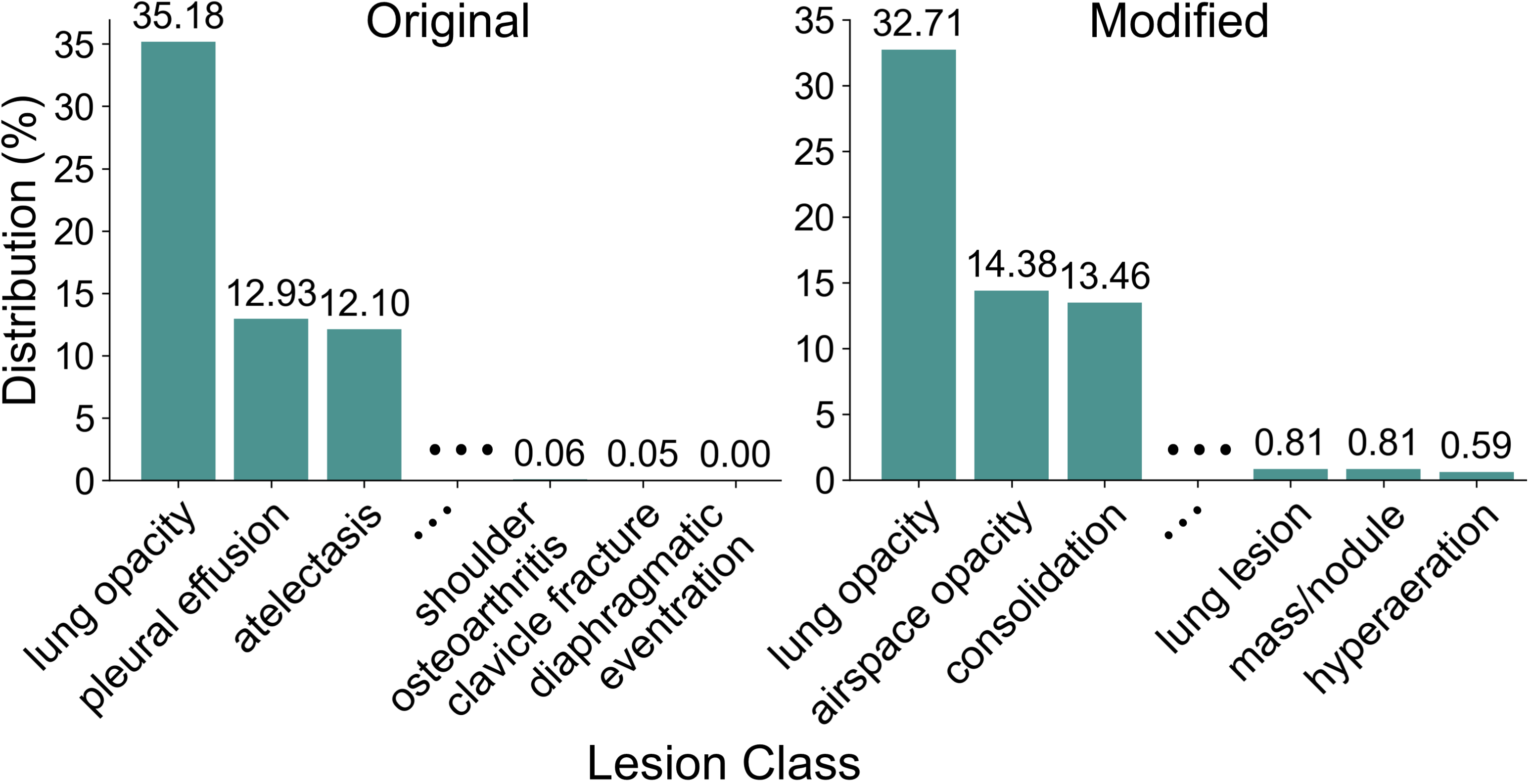}
\caption{Distribution of the lesion classes showing only the first three and last three classes. Original (42 classes); Modified (21 classes).}
  \label{fig-long-tail-modified}
\end{figure}

The proposed lesion detector is implemented on the Chest ImaGenome Dataset \cite{chestIma}, which provides coordinates and lesion labels for 42 classes across anatomical regions in CXR images. However, as shown on the left of Fig. \ref{fig-long-tail-modified}, the dataset suffers from a highly imbalanced distribution, which can significantly impair computer vision tasks \cite{long-tail-survey}.

\begin{figure}[!htbp]
  \centering
  \includegraphics[draft=false, width=\columnwidth]{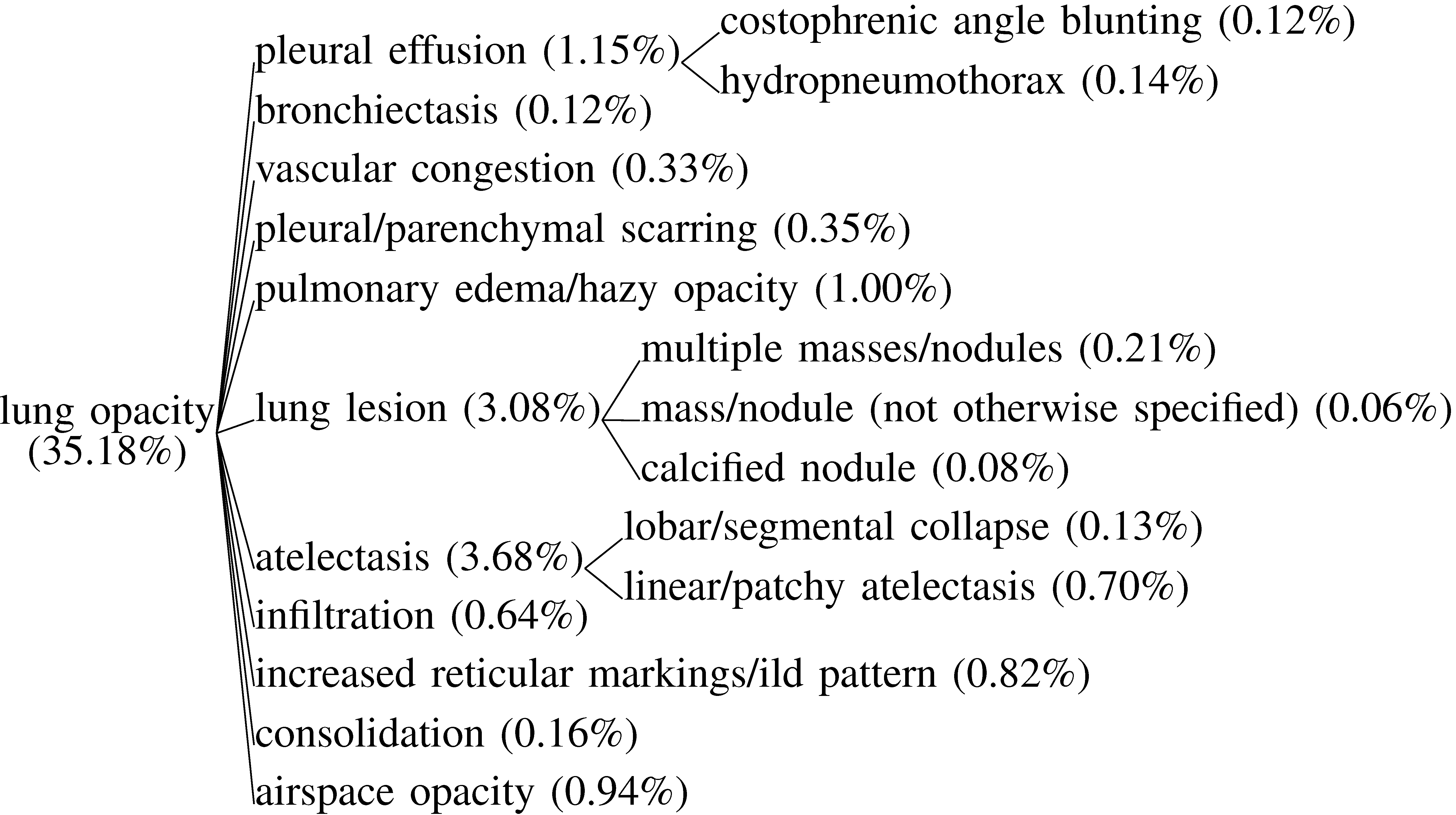}
\caption{Parent-child relationship in the Chest ImaGenome dataset  \protect\cite{chestIma}, with 'Lung opacity' as the root node. It branches into second-level nodes like 'pleural effusion' and 'lung lesion,' which further split into third-level nodes. The proportion of each node among the 42 lesion classes is indicated in brackets.}
    \label{fig:parent-child}
\end{figure}

To address the long-tail effect and prioritize crucial lesion classes, we initially eliminate tail classes that constitute less than 0.5\% of the training data. Furthermore, as shown in Fig. \ref{fig:parent-child}, lesion classes display a hierarchical parent-child relationship that leads to redundancies. For example, identifying a child lesion node within an anatomical region automatically includes all corresponding parent lesion nodes in the class vector $c_i$, introducing redundancy that potentially biases object detection towards these less specific parent classes. Notably, 'lung opacity' is the only class with child nodes and is disproportionately represented. To mitigate this bias, we remove the root node 'lung opacity' from the class vector $c_i$ in the image label and retain all other associated nodes whenever a third-level node is present. This modification, shown in Fig. \ref{fig-long-tail-modified}, reduces the classes to 21, effectively minimizing redundancy and correcting the imbalance in lesion class distribution.

\subsection{Pathology-Aware Regional Prompts}

To emulate the working pattern of radiologists and explicitly guide the report decoder with diverse anatomical and pathological information, we introduce pathology-aware regional prompts. These prompts consist of 29 tokens, each representing a single lesion finding in one of the 29 anatomical regions. The prompt $P \in \mathbb{R}^{29 \times 1}$ is structured as a string composed of 29 tokens $\{R_1, R_2, \ldots, R_{29}\}$, where $R_i$ represents the lesion token for the $i_{th}$ region. The construction of these prompts during the training and inference phases is detailed in the following sections.

\subsubsection{Training}

To effectively enhance report generation with prompt guidance, we prefix ground truth prompt labels $P$ to reference reports, facilitating the model's learning by highlighting the link between the prompt and the report. These labels originate from scene graphs that provide the locations and lesion findings of each anatomical region.

To simplify and ensure clarity, we employ a rule-based strategy for constructing region-level prompt labels. As outlined in Section \ref{class-reduction}, modifications are required only for 'lung opacity' and its child nodes, as other lesions do not have child nodes. The rules for organizing labels for each anatomical region are as follows:

\begin{enumerate} 
    \item If a region's label does not have child nodes or contains only 'lung opacity', it is assigned its own token.
    \item If a second-level node accompanies 'lung opacity', we use the token for that second-level lesion, as it is more diagnostically specific and informative than the root node.
    \item If a third-level node is present, indicating the presence of its two parent nodes, we continue to use the second-level node's token. This approach balances specificity with practicality. The third-level node, while precise, is typically under-represented in the dataset, making it less likely to be detected and potentially causing discrepancies between training and inference. The second-level node, though less explicit, is detected more frequently and provides reliable diagnostic guidance.
    \item Finally, if multiple lesions remain after applying these rules, the least common disease is retained to further mitigate the long-tail effect.
\end{enumerate}

 This systematic labeling strategy ensures that each anatomical region is associated with the most indicative and distinct disease token, facilitating precise and informative training of the report decoder.

\subsubsection{Inference}

Upon receiving an input image, anatomical region detection and lesion detection are initially performed. Detected lesions are then assigned to overlapping anatomical regions based on the IoU of their locations, as illustrated in the centre of Fig \ref{fig1-pipeline}. Specifically, a lesion bbox is assigned to an anatomical region if their IoU exceeds a predefined threshold. If multiple lesion bboxes overlap with a single anatomical region, only the bbox with the highest IoU is considered. Within that chosen bbox, we employ the same rule-based construction strategy introduced in training. Subsequently, each anatomical region is assigned a corresponding token $R_i$ that represents the identified lesion.

If an anatomical region falls below the IoU threshold or goes undetected, the corresponding token is set to '[NEG]'. After assigning tokens to all 29 regions, a pathology-aware regional prompt is constructed.

\subsection{Report Decoder}

This component functions as the language decoder, utilizing the anatomy-level visual features $V_a \in \mathbb{R}^{29 \times 1024}$  where each region is represented by a 1024-dimensional vector, and the pathology-aware regional prompts $P \in \mathbb{R}^{29 \times 1}$, to generate a diagnostic report. We employ the BERT-base model \cite{BERT} configured in decoder mode without pre-training. The BERT model, widely adopted for its effectiveness in generating chest X-ray reports, excels in processing inputs from both visual and linguistic domains. Its self-attention mechanism adeptly integrates cross-modal inputs by focusing on the most pertinent aspects of the data, whether they are image regions or textual cues. The process of decoding is formulated as:
\begin{equation}
    Report = BERT(V_a; R_1, R_2, \ldots , R_{29}).
\end{equation}

During training, pathology prompts are prefixed to the reference report and inputted into the BERT decoder post-tokenization. In inference mode, only the pathology prompts are supplied as textual guidance, effectively serving as a trigger for the decoder to commence report generation. 

\section{Experiments and Discussions}
\subsection{Datasets}
In this study, we utilize the Chest ImaGenome v1.0.0 dataset \cite{chestIma}, which is derived from the widely used MIMIC-CXR-JPG dataset \cite{johnson2019mimic, johnson2019mimic-jpg} comprising image-report pairs from approximately 65,000 patients. This dataset enriches the MIMIC dataset with an automatically constructed scene graph for each frontal chest X-ray image, providing bounding box coordinates for 29 distinct anatomical regions and multiple disease findings per region. We follow the official dataset split and discard samples lacking the \textit{Findings} section, resulting in 113,915 training samples, 15,658 validation samples, and 32,711 test samples.

While training the multi-label lesion detector, we follow the YOLOv5 official documentation \cite{ultralytics2021yolov5} to filter out all but 12\% of the negative images and modify the lesion labels as described in Section \ref{class-reduction}.

\subsection{Implementation Details}
Our pipeline is trained in two stages using PyTorch 2.0 on a single NVIDIA RTX 4090 GPU.

\subsubsection{Anatomical Region Detector} Images are resized to 512 pixels on the shorter side while maintaining aspect ratio, and cropped to 512$\times$512. We employ random cropping during training and centre cropping for inference. The learning rate is initially set at 1e-3 and adjusted using an AdamW optimizer \cite{adamw} with a weight decay of 1e-2. Training runs for 10 epochs with a batch size of 16.

\subsubsection{Multi-Label Lesion Detector} Images are resized to 640$\times$640, with the model running for 80 epochs at a batch size of 50. The confidence threshold is set at 0.35 and the IoU threshold at 0.45 for non-maximum suppression during inference. $L_{\text{cls}}$, $L_{\text{obj}}$ and $L_{\text{box}}$ are set at 0.5, 1.0, and 0.05, respectively.

\subsubsection{Full Model} The IoU threshold for assigning lesion findings to anatomical regions is set to 0.4. The hidden size of the BERT model is set to 1024, consistent with the dimension of the anatomy-level visual features $V_a$. An AdamW optimizer with a weight decay of 0.05 is used, and the initial learning rate (LR) of 5e-5 is set following a cosine schedule. A beam search with a width of 4 is employed for sentence generation, with training conducted over 15 epochs at a batch size of 14.

\subsection{Evaluation Metrics}
\subsubsection{Automated Metrics}
We evaluate the report generation performance of our model by the popular Natural Language Generation (NLG) and clinical efficacy (CE) metrics. NLG metrics are crucial for evaluating the textual similarity between the generated text and the ground truth. In this study, we report the BLEU-n \cite{papineni2002bleu}, METEOR\cite{banerjee2005meteor} and ROUGE-L \cite{lin2004rouge}, which measure text quality by counting matching n-grams (i.e., word overlap). However, NLG metrics can fall short in radiology, sometimes assigning high scores to descriptions with opposing medical meanings. Therefore, CE metrics that prioritize the accuracy of medical observations over text fluency or grammar are critical for assessing the model's ability to capture essential clinical findings. In this study, we utilize Precision, Recall, and F1-score based on 14 common medical observations extracted using the CheXbert labeler \cite{smit2020chexbert}.

\subsubsection{Formal Expert Evaluation}
Evaluating radiology reports is challenging due to the diverse reporting styles of radiologists. Automated metrics like NLG and CE offer valuable quantitative insights into the quality of generated reports. NLG metrics excel in assessing textual similarity, while CE metrics effectively evaluate the presence of critical pathologies. However, to fully capture the nuances of clinical language and judgment, particularly the appropriateness of key details and the potential impact of redundant information, expert evaluation remains essential. Therefore, we collaborated with certified clinical experts (M. Komorowski and D. Marshall) to develop more relevant clinical metrics. In this study, clinicians evaluated responses from 100 random samples in our test set. To ensure impartiality and minimize bias, responses from all models were randomized and anonymized. The metrics are defined as follows:

\begin{enumerate}[leftmargin=*,align=left]
    \item \textbf{Rubric}: Following Yang et al. \cite{rubric}, clinicians compare the quality of two radiology reports by grading from X, B2, B1, C, A2 and A1 (maps to 1-5 in our experiments).
    \item \textbf{Brevity}: Clinicians assess report verbosity by: Too Concise (-1), Good (0), and Too Verbose (+1).
    \item \textbf{Accuracy (1-5)}: As defined in Table \ref{tab:accuracy}, clinicians evaluate report quality in the absence of a good reference, focusing on disease identification, detail of findings, and impact on patient management \cite{sharma2024}.
    \item \textbf{Danger (0/1)}: Clinicians assess whether a report indicates an immediate severe health risk (Yes or No) to prevent potential harm from real clinical applications.
\end{enumerate}

\subsubsection{Object Detection}
We assess the performance of the anatomical region detector using the Intersection over Union (IoU) metric, calculated as $\text{IoU} = \frac{\text{Area of Intersection}}{\text{Area of Union}}$, and report the average number of detected regions per image.

For the multi-label lesion detector, we evaluate using standard object detection metrics: Precision, Recall, and mean Average Precision (mAP) at IoU thresholds of 0.5 and 0.95.

\begin{table}[htbp]
\caption{Accuracy Definition in Formal Expert Evaluation  \protect\cite{sharma2024}}
\label{tab:accuracy}
\centering
\small
\renewcommand{\arraystretch}{1.2}
\resizebox{\columnwidth}{!}{
\begin{tabular}{cl}
\Xhline{1pt}
\multicolumn{1}{l}{Score} & \multicolumn{1}{c}{Definition}        \\ \Xhline{1pt}
5                                  & Perfect report, accurately detailed (no hallucinations) \\ \hline
4                                  & Generally accurate, a few missing details             \\ \hline
3 &
  \begin{tabular}[c]{@{}l@{}}Key details present but require additional interpretation \\[-0.8ex] with no issues regarding patient management \end{tabular} \\ \hline
2 &
  \begin{tabular}[c]{@{}l@{}}Missing key details but not dangerous \end{tabular} \\ \hline
1                                  & Dangerous (would lead to mismanagement)        \\ \Xhline{1pt}      
\end{tabular}
}
\end{table}

\vspace{-5pt}

\subsection{Automated Metrics Comparison}

\begin{table*}[htbp]
\caption{Comparison with state-of-the-art methods on the MIMIC-CXR dataset using NLG and CE metrics. For LLM-based methods, the number of parameters is indicated in billions (B). The performance of XrayGPT are referenced from  \protect\cite{liu2024bootstrapping}. * denotes methods re-implemented and tested by us following official implementations. All other results are sourced from original publications. BL, MTR, and RG-L represent BLEU, METEOR, and ROUGE-L, while P, R, and F1 denote Precision, Recall, and F1-score, respectively.}
\label{tab:comparison}
\centering
\small
\renewcommand{\arraystretch}{1.2}
\begin{tabular}{lclcccccclccc}
\Xhline{1pt}
\multicolumn{1}{l}{\multirow{2}{*}{Model}} &
  \multicolumn{1}{l}{\multirow{2}{*}{Year}} &
  \multirow{2}{*}{LLM} &
  \multicolumn{6}{c}{NLG metrics} &
   &
  \multicolumn{3}{c}{CE metrics} \\ \cline{4-9} \cline{11-13} 
\multicolumn{1}{c}{} &
  \multicolumn{1}{l}{} &
   &
  BL-1 &
  BL-2 &
  BL-3 &
  BL-4 &
  MTR &
  RG-L &
   &
  P &
  R &
  F1 \\ \Xhline{1pt}
R2Gen \cite{r2gen} &
  2020 &
  \xmark &
  0.353 &
  0.218 &
  0.145 &
  0.103 &
  0.142 &
  0.277 &
   &
  0.333 &
  0.273 &
  0.276 \\
R2GenCMN \cite{r2gencmn} &
  2021 &
  \xmark &
  0.353 &
  0.218 &
  0.148 &
  0.106 &
  0.142 &
  0.278 &
   &
  0.334 &
  0.275 &
  0.278 \\
ClinicalBERT \cite{clinical-bert} &
  2022 &
  \xmark &
  0.383 &
  0.230 &
  0.151 &
  0.106 &
  0.144 &
  0.275 &
   &
  0.397 &
  0.435 &
  0.415 \\
METrans \cite{METransformer} &
  2023 &
  \xmark &
  0.386 &
  0.250 &
  0.169 &
  0.124 &
  0.152 &
  0.291 &
   &
  0.364 &
  0.309 &
  0.311 \\
XrayGPT \cite{XrayGPT} &
  2023 &
  7B &
  0.128 &
  0.045 &
  0.014 &
  0.004 &
  0.079 &
  0.111 &
   &
  - &
  - &
  0.326 \\
CAMANet \cite{CAMANet} &
  2023 &
  \xmark &
  0.374 &
  0.230 &
  0.155 &
  0.112 &
  0.145 &
  0.279 &
   &
  0.483 &
  0.323 &
  0.387 \\
RGRG* \cite{tanida2023interactive} &
  2023 &
  \xmark &
  0.373 &
  0.249 &
  \textbf{0.175} &
  \textbf{0.126} &
  0.168 &
  0.264 &
   &
  0.461 &
  \textbf{0.475} &
  0.447 \\
\begin{tabular}[c]{@{}l@{}}AdaMatch\\[-0.8ex] -Cyclic \cite{adamatch}\end{tabular} &
  2024 &
  3B &
  0.379 &
  0.235 &
  0.154 &
  0.106 &
  0.163 &
  0.286 &
   &
  - &
  - &
  - \\
MedDr \cite{meddr}&
  2024 &
  34B &
  0.322 &
  - &
  - &
  0.072 &
  \textbf{0.238} &
  0.226 &
   &
  - &
  - &
  - \\
CheXagent* \cite{chexagent}&
  2024 &
  7B &
  0.200 &
  0.123 &
  0.083 &
  0.058 &
  0.104 &
  0.249 &
   &
  0.477 &
  0.273 &
  0.348 \\
PromptMRG* \cite{jin2023promptmrg}&
  2024 &
  \xmark &
  0.387 &
  0.230 &
  0.147 &
  0.100 &
  0.148 &
  0.260 &
   &
  0.505 &
  0.461 &
  0.453 \\ \hline
Ours &
  2024 &
  \xmark &
  \textbf{0.394} &
  \textbf{0.251} &
  0.173 &
  \textbf{0.126} &
  0.151 &
  \textbf{0.302} &
   &
  \textbf{0.509} &
  0.437 &
  \textbf{0.470} \\ \Xhline{1pt}
\end{tabular}
\end{table*}

We evaluate the performance of our model against various state-of-the-art (SOTA) baseline methods on automated metrics. These methods include R2Gen \cite{r2gen}, R2GenCMN \cite{r2gencmn}, Clinical-BERT \cite{clinical-bert}, METrans \cite{METransformer}, CAMANet \cite{CAMANet}, RGRG \cite{tanida2023interactive}, PromptMRG \cite{jin2023promptmrg}, and VLM-based approaches such as XrayGPT \cite{XrayGPT}, MedDr \cite{meddr}, CheXagent \cite{chexagent} and AdAMatch-Cyclic \cite{adamatch}. These approaches incorporate a variety of techniques ranging from conventional encoder-decoder architectures, finetuning pre-trained BERT models, and aligning cross-modal features, to employing retrieval augmentation and large-scale vision-language models (VLMs). This diverse and comprehensive comparison underscores the effectiveness and rationality of our proposed model.

Table \ref{tab:comparison} demonstrates that our model excels in radiology report generation, outperforming previous SOTA baseline models in 4 out of 6 NLG metrics and 2 out of 3 CE metrics. Notably, it shows the highest difference in BLEU-1, ROUGE-L, Precision and F1-score.

Notably, RGRG achieves the second-best performance overall by providing sentence-level descriptions for each anatomical region separately. In contrast, our model utilizes anatomy-level visual features of various scales and explicitly leverages global pathological information through pathology-aware regional prompts. This strategy leads to improvements across most metrics, including a notable 2.3\% increase in F1-score, underscoring our model's enhanced clinical accuracy. The comparison with RGRG demonstrates the significant benefits of aligning anatomical and pathological information in the report generation process.

Furthermore, PromptMRG employs a disease classification branch with cross-modal feature enhancement, achieving robust overall performance. However, it is limited to 15 disease classes and lacks disease localization capabilities. Conversely, our model accurately identifies 21 lesions and correlates them with specific anatomical regions under different scales, surpassing the patch-level capabilities of Vision Transformer used in PromptMRG. This approach yields improvements across all metrics, including a significant 1.7\% increase in F1-score and the highest Precision with minimal false positives. Such precision is crucial for avoiding misdiagnosis and enhancing treatment planning, underscoring the advantages of region-level pathology localization.

Table \ref{tab:comparison} also reveals that VLMs generally underperform on NLG metrics compared to non-LLM methods, notably XrayGPT, which exhibits the lowest scores. This likely stems from several factors: LLMs are typically pre-trained on diverse linguistic datasets and later fine-tuned on radiology-specific data. The scarcity of high-quality CXR image-report pairs hampers the training of large LLMs, resulting in lower textual similarity. In contrast, traditional methods leverage specialized medical knowledge bases and tailored workflows, with often smaller language decoders and being trained from scratch on targeted datasets. This approach allows more controlled training and easier optimization for textual similarity. Notably, the AdAMatch-Cyclic strategy with the smaller dolly-v2-3b \cite{dollyv3} outperforms other LLMs, highlighting the impact of limited training data on larger models.

\begin{table}[htbp]
\centering
\small
\renewcommand{\arraystretch}{1.2}
\caption{Comparison of formal expert evaluation results.}
\label{tab:clinical-performance}
\begin{tabular}{lcccc}
\Xhline{1pt}
\multicolumn{1}{c}{Model} &
  \begin{tabular}[c]{@{}c@{}}Rubric\\[-0.6ex] $\uparrow$\end{tabular} &
  \begin{tabular}[c]{@{}c@{}}Brevity\\[-0.6ex] $\bigtriangleup\downarrow$\end{tabular} &
  \begin{tabular}[c]{@{}c@{}}Accuracy\\[-0.6ex] $\uparrow$ \end{tabular} &
  \begin{tabular}[c]{@{}c@{}}Danger\\[-0.6ex] $\downarrow$\end{tabular} \\ \Xhline{1pt}
CheXagent \cite{chexagent} & 2.28 & 0.08 & \textbf{3.55} & 0.03 \\
PromptMRG \cite{jin2023promptmrg} & \textbf{2.32} & 0.19 & 3.49 & 0.03 \\
RGRG \cite{tanida2023interactive}      & 2.21 & 0.40 & 3.23 & 0.14 \\ \hline
Ours      & 2.26 & \textbf{0.01} & 3.51 & \textbf{0.03} \\ \Xhline{1pt}
\end{tabular}%
\end{table}

\subsection{Formal Expert Evaluation}

In the expert evaluation, our model is compared against three state-of-the-art baselines: RGRG \cite{tanida2023interactive}, PromptMRG \cite{jin2023promptmrg}, and CheXagent \cite{chexagent}. The first two models employ enhanced encoder-decoder architectures, while CheXagent is a large-scale Vision-Language Model (VLM).

The results of the four models on expert metrics are shown in Table \ref{tab:clinical-performance}. Our model achieves unparalleled conciseness with the best Brevity score of 0.01 and excels in minimizing the risk of critical errors with the lowest Danger score of 0.03. It also performs comparably well in Rubric and Accuracy, closely aligning with reference reports and effectively capturing essential radiological details, demonstrating a well-rounded superiority in clinical evaluations.

Notably, CheXagent, though achieving the highest Accuracy of 3.55, scores moderately on Brevity, suggesting its reports may be detailed but potentially verbose. PromptMRG shows robust adherence to clinical guidelines with a slightly higher Rubric score than our model but falls short in Brevity and safety, indicating areas for potential refinement. RGRG, while achieving the second-best performance in automated metrics evaluation, performs moderately overall, as indicated by the highest Danger score. This suggests that a balance between quantitative success and clinical safety still needs to be achieved.

\begin{figure*}[htbp]
  \centering
  \includegraphics[draft=false, width=0.85\textwidth]{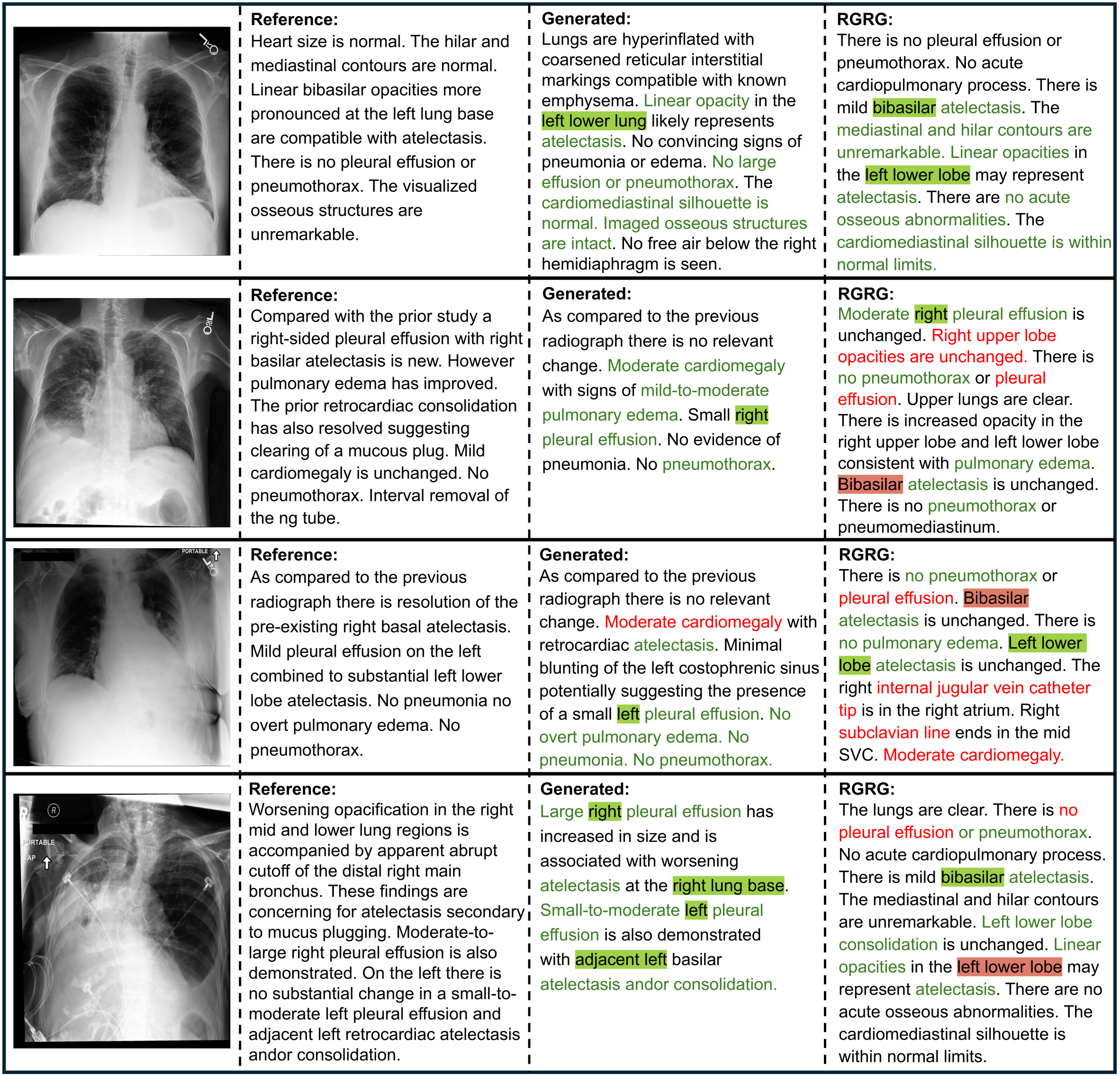}
\caption{Examples of generated reports by our model and RGRG  \protect\cite{tanida2023interactive}. Green font indicates descriptions consistent with the reference report, while red font denotes incorrect descriptions of negative or unmentioned pathologies. Green highlights confirm correct pathology locations, and red highlights point out error locations.}
  \label{fig:qualitative-1}
\end{figure*}

By synthesizing insights from both quantitative metrics in \ref{tab:comparison} and expert evaluations, it is evident that our model not only adheres to clinical reporting standards by closely aligning with expert assessments and rigorously upholding medical guidelines, but also surpasses other models in key automated metrics. This superior performance highlights its effectiveness and reliability, and potential to streamline diagnostic processes and enhance the clinical reporting workflow.

\subsection{Instance Performance}

Several report examples in Fig. \ref{fig:qualitative-1} compare our model's responses to RGRG \cite{tanida2023interactive}, which achieves second-best in automated metrics comparison. It also uses anatomical regions but treats them separately and does not include other diagnostic information. These examples show that our model consistently identifies key pathologies and produces diagnostic reports closely aligned with the ground truth, potentially aiding radiology examinations and improving clinical efficiency.

Moreover, our model accurately identifies the presence, extent, and location of pathologies. As highlighted in the last row of Fig. \ref{fig:qualitative-1}, it precisely reports a large right-sided and a small-to-moderate left-sided pleural effusion. In contrast, RGRG overlooks these effusions and incorrectly reports linear opacities in the opposite lung, likely due to the mirror image effect in radiological imaging that reverses laterality. Additionally, it hallucinates several pathologies, particularly regarding medical devices, as illustrated in the third row. Our model addresses these issues by effectively integrating anatomical and pathological information through the prompt guidance, ensuring precise localization and description.

\begin{table*}[htbp]
\centering
\small
\caption{Ablation study comparing baseline and enhanced models using text prompts and pathology-aware regional prompts (PARP).}
\label{tab:ablation}
\renewcommand{\arraystretch}{1.2}
\begin{tabular}{lcllllllcll}
\Xhline{1pt}
\multicolumn{1}{c}{\multirow{2}{*}{Model}} & \multicolumn{6}{c}{NLG metrics}                                            &  & \multicolumn{3}{c}{CE metrics}            \\ \cline{2-7} \cline{9-11} 
\multicolumn{1}{c}{} &
  BL-1 &
  \multicolumn{1}{c}{BL-2} &
  \multicolumn{1}{c}{BL-3} &
  \multicolumn{1}{c}{BL-4} &
  \multicolumn{1}{c}{MTR} &
  \multicolumn{1}{c}{RG-L} &
   &
  P &
  \multicolumn{1}{c}{R} &
  \multicolumn{1}{c}{F1} \\ \Xhline{1pt}
Baseline                         & \multicolumn{1}{l}{0.364} & 0.237 & 0.166 & 0.122 & 0.141 & 0.300 &  & \multicolumn{1}{l}{0.464} & 0.323 & 0.381 \\
+Text Prompt                                & \multicolumn{1}{l}{0.261} & 0.134 & 0.081 & 0.053 & 0.112 & 0.274          &  & \multicolumn{1}{l}{0.397} & 0.196 & 0.263 \\ \hline
\textbf{+PARP (Ours)} &
  \textbf{0.394} &
  \multicolumn{1}{c}{\textbf{0.251}} &
  \multicolumn{1}{c}{\textbf{0.173}} &
  \multicolumn{1}{c}{\textbf{0.126}} &
  \multicolumn{1}{c}{\textbf{0.151}} &
  \multicolumn{1}{c}{\textbf{0.302}} &
   &
  \textbf{0.509} &
  \multicolumn{1}{c}{\textbf{0.437}} &
  \multicolumn{1}{c}{\textbf{0.470}} \\ \Xhline{1pt}
\end{tabular}
\end{table*}

Remarkably, RGRG sometimes produces contradictory statements. For instance, in the second row, it correctly describes a moderate right pleural effusion but then denies any pleural effusion in the image. This inconsistency arises from providing separate, sentence-level descriptions for each region without considering their potential overlaps or connections. In contrast, our model integrates anatomy-level visual features and pathological information at various scales, enabling effective comparison and enhancing report consistency.

The instance performance of our model underscores its ability to handle complex radiology diagnostic tasks, showcasing its strengths in comprehensive radiology report generation.

\subsection{Ablation Study}

To verify the effectiveness of our pathology-aware regional prompts, we conduct an ablation study by removing the lesion detector and prompt guidance. In this baseline configuration, the report decoder is activated by a predefined starting token and utilizes solely the anatomy-level visual features.

Additionally, we design a text-prompting strategy that encompasses all detected lesions across every anatomical regions in the following manner:

\textit{"Please generate a report for this chest x-ray image. Here are some initial findings: "}
\begin{enumerate}[leftmargin=*,align=left]
    \item \textit{\textbf{[Lesion names]} may be present in the \textbf{[Region names]}. }
    \item \ldots
\end{enumerate}

In this setup, \textit{Lesion names} and \textit{Region names} are presented as lists to account for potential overlaps of lesions across different regions. These prompts are only utilized during inference due to their variable length. This method tests whether detailed, previously unseen text prompts can enhance report generation during testing.

As shown in Table \ref{tab:ablation}, ablation studies reveal significant improvements in all NLG and CE metrics after integrating pathology-aware regional prompts, which explicitly provide anatomical and pathological information. Specifically, the F1-score increases by 8.9\%, highlighting the crucial role of these prompts in producing diagnostically accurate reports. In contrast, the text prompt strategy, despite being descriptive, yields the poorest results across all metrics.

\begin{figure}[htbp]
  \centering
  \includegraphics[draft=false, width=\columnwidth]{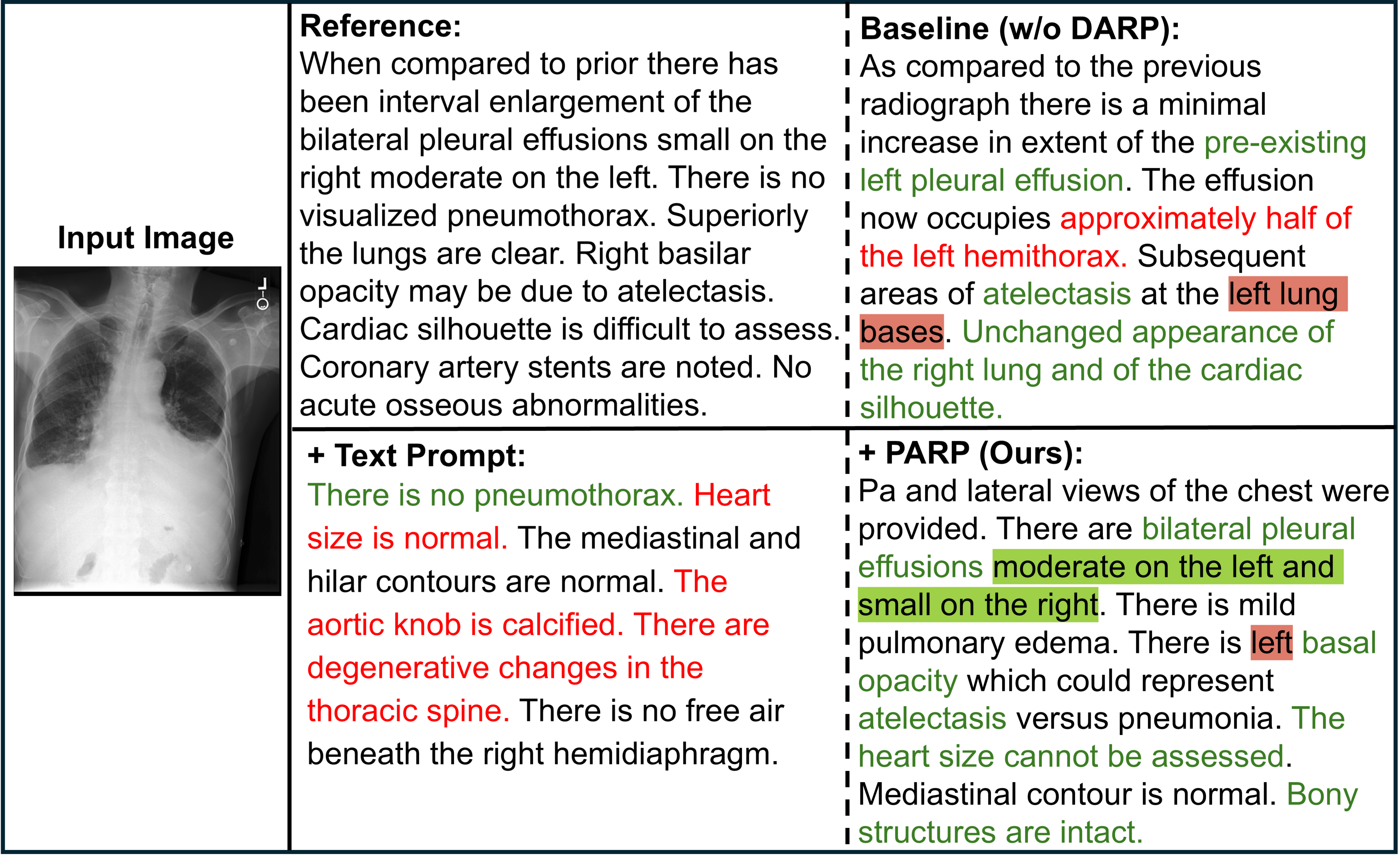}
\caption{Examples of the ablation study. Green font indicates descriptions consistent with the reference report, while red font denotes incorrect descriptions of negative or unmentioned pathologies. Green highlights confirm correct pathology locations, and red highlights point out error locations.}
  \label{fig:ablation}
\end{figure}

Moreover, Fig. \ref{fig:ablation} presents examples from the three methods, aligning well with their quantitative performances. The baseline model identifies some key findings like left pleural effusion but misreports its extent. Incorporating pathology-aware regional prompts, our model accurately detects and describes crucial diagnostic details, correctly noting pleural effusions as moderate on the left and small on the right without false positives. This alignment highlights the effectiveness of our methodological adjustments.

Conversely, the text prompt method performs poorly, often misdiagnosing diseases and generating many false positives. This suggests that verbose prompts during inference hinder performance, emphasizing the need for concise prompts to effectively guide the report decoder.

\subsection{Object Detectors Performance}
\subsubsection{Anatomical Region Detector}

Table \ref{tab:anatomy-detector-short} shows the micro-average IoU scores for six primary regions. Our model attains an average IoU of 0.892 across 29 regions, highlighting its comprehensive accuracy. Moreover, it consistently identifies an average of 28.943 regions per image, highlighting its precision in anatomical identification. We believe this effectiveness is largely due to the well-distributed anatomical regions per image, which optimize model training.

\begin{table}[htbp]
\centering
\caption{Micro average IoU of the anatomical region detector across six primary regions and overall average for all 29 regions.}
\label{tab:anatomy-detector-short}
\renewcommand{\arraystretch}{1.2}
\resizebox{\columnwidth}{!}{
\begin{tabular}{ccccccc|c}
\hline
Region &
  \begin{tabular}[c]{@{}c@{}}Right\\[-0.7ex] Lung\end{tabular} &
  \begin{tabular}[c]{@{}c@{}}Left\\[-0.7ex] Lung\end{tabular} &
  Spine &
  Abdomen &
  \begin{tabular}[c]{@{}c@{}}Left \\[-0.7ex] Mid Lung\end{tabular} &
  \begin{tabular}[c]{@{}c@{}}Media-\\[-0.7ex] stinum\end{tabular} &
  \textbf{Avg.} \\ \hline
IoU &
  0.929 &
  0.925 &
  0.944 &
  0.923 &
  0.900 &
  0.875 &
  \textbf{0.892} \\ \hline
\end{tabular}%
}
\end{table}

\subsubsection{Multi-Label Lesion Detector}

Table \ref{tab:yolo-performance} shows the lesion detector's performance on conventional object detection metrics. To our best knowledge, this is the first work that utilizes the Chest ImaGenome dataset to perform multi-label lesion detection, with no existing benchmarks for comparison. Our model demonstrates robust performance, achieving an overall average precision of 45.4\% and a mAP@0.5 of 0.345, effectively identifying all targeted pathologies. Moreover, our model successfully addresses the persistent long-tail effect, with the most frequent condition 'lung opacity' not exhibiting overfitting, and achieving precision and recall above 0.57. Other rarer conditions like hyperaeration and pneumothorax also demonstrate excellent performance.

\begin{figure*}[!ht]
  \centering
  \includegraphics[draft=false, width=\textwidth]{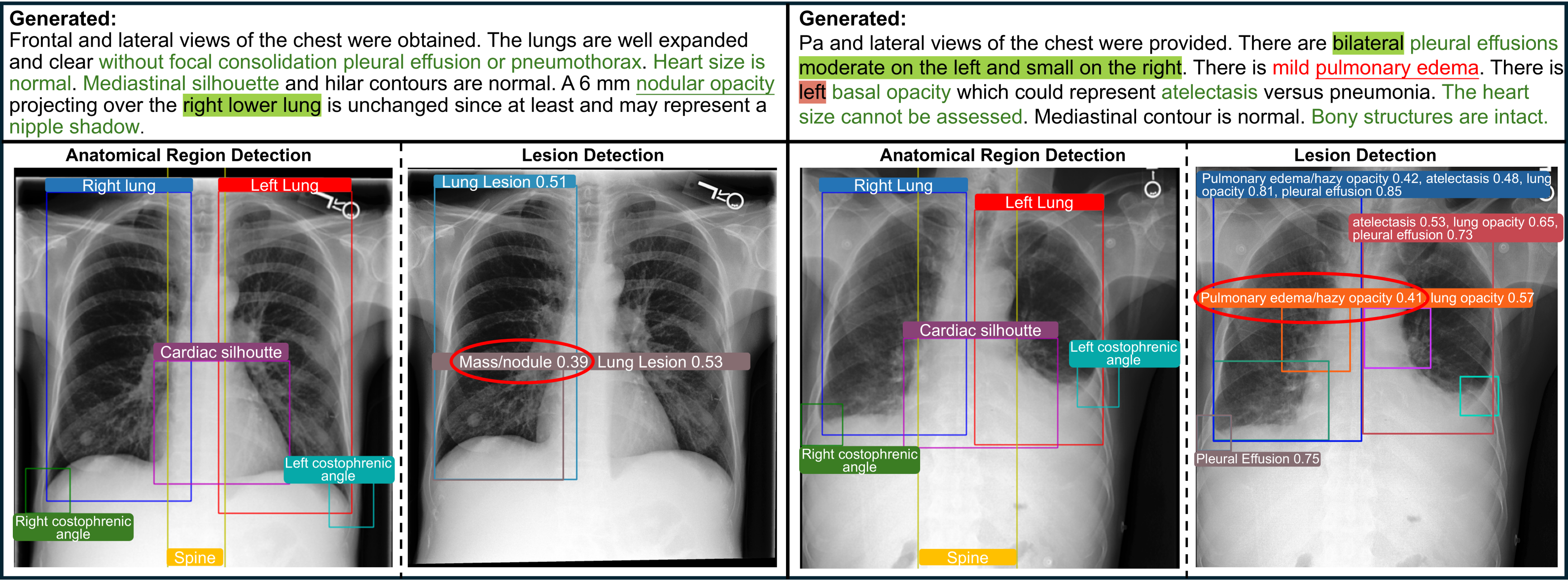}
\caption{Examples of detection results and generated reports. Each image includes detections across 29 anatomical regions, with only the first 6 shown for clarity. Lesion bounding boxes are labeled with detected lesions and their confidence scores, with some labels hidden in the right example for conciseness. Underlines and circles highlight examples of highly aligned findings between the report and detected lesions.}
  \label{fig:more-example}
\end{figure*}

\begin{table}[!htbp]
\centering
\small
\caption{Lesion detector performance on standard object detection metrics. The underlined text highlights second-level lesions defined in Fig. \ref{fig:parent-child}, which are prioritized in constructing prompts. P and R denote Precision and Recall.}
\label{tab:yolo-performance}
\renewcommand{\arraystretch}{1.1}
\resizebox{\columnwidth}{!}{
\begin{tabular}{l|c|c|c|c}
\Xhline{1pt}
Lesion                                                                & P      & R         & mAP@0.5       & mAP@0.95      \\ \Xhline{1pt}
lung opacity                                                           & 0.575          & 0.646          & 0.577          & 0.451          \\
airspace opacity                                                       & {\ul 0.406}    & {\ul 0.061}    & {\ul 0.222}    & {\ul 0.191}    \\
consolidation                                                          & {\ul 0.352}    & {\ul 0.075}    & {\ul 0.203}    & {\ul 0.176}    \\
atelectasis                                                      & {\ul 0.561}    & {\ul 0.516}    & {\ul 0.5}      & {\ul 0.412}    \\
\begin{tabular}[c]{@{}l@{}}linear/patchy\\[-0.8ex]  atelectasis\end{tabular}                                               & 0.351          & 0.121          & 0.207          & 0.195          \\
\begin{tabular}[c]{@{}l@{}}lobar/segmental/\\[-0.8ex] collapse\end{tabular}                                                & 0.376          & 0.054          & 0.21           & 0.154          \\
\begin{tabular}[c]{@{}l@{}}pulmonary edema/\\[-0.8ex] hazy opacity\end{tabular}          & {\ul 0.493} & {\ul 0.549} & {\ul 0.478} & {\ul 0.369} \\
vascular congestion                                                    & {\ul 0.406}    & {\ul 0.328}    & {\ul 0.294}    & {\ul 0.244}    \\
vascular redistribution                                                & 0.5            & 0.01           & 0.253          & 0.24           \\
pleural effusion                                                       & {\ul 0.589}    & {\ul 0.531}    & {\ul 0.558}    & {\ul 0.401}    \\
\begin{tabular}[c]{@{}l@{}}costophrenic angle\\[-0.8ex] blunting\end{tabular}  & 0.394          & 0.047          & 0.214          & 0.184          \\
\begin{tabular}[c]{@{}l@{}}pleural/parenchymal\\[-0.8ex] scarring\end{tabular}           & {\ul 0.431} & {\ul 0.187} & {\ul 0.277} & {\ul 0.242} \\
\begin{tabular}[c]{@{}l@{}}enlarged cardiac \\[-0.8ex] silhouette\end{tabular} & 0.608          & 0.757          & 0.681          & 0.505          \\
mediastinal widening                                                   & 0.349          & 0.092          & 0.196          & 0.153          \\
enlarged hilum                                                         & 0.387          & 0.102          & 0.224          & 0.19           \\
tortuous aorta                                                         & 0.442          & 0.46           & 0.39           & 0.326          \\
vascular calcification                                                 & 0.439          & 0.416          & 0.362          & 0.299          \\
pneumothorax                                                           & 0.606          & 0.335          & 0.46           & 0.39           \\
 lung lesion                                                    & {\ul 0.41}     & {\ul 0.174}    & {\ul 0.272}    & {\ul 0.242}    \\
\begin{tabular}[c]{@{}l@{}}mass/nodule\end{tabular} & 0.349       & 0.124       & 0.216       & 0.187       \\
hyperaeration                                                          & 0.512          & 0.418          & 0.453          & 0.435          \\ \hline
\textbf{Average}                                                           & \textbf{0.454} & \textbf{0.285} & \textbf{0.345} & \textbf{0.285} \\ \Xhline{1pt}
\end{tabular}
}
\end{table}

Despite achieving high precision and mAP for second-level nodes, the recall is generally lower, particularly for prevalent conditions like airspace opacity and consolidation where recall is below 0.1. This is likely due to the high prevalence of 'lung opacity' and its similarity to these conditions, complicating the differentiation and lowering recall for less distinctive diseases. In contrast, conditions unrelated to 'lung opacity', such as hyperaeration (0.59\%) and pneumothorax (0.82\%), achieve excellent results despite their rarity. Nonetheless, this challenge is unlikely to adversely affect report generation, as the observed detection patterns reflect actual reports, where lung opacity often coexists with these secondary conditions. Therefore, even with occasional omissions of secondary nodes in the prompts, leveraging lung opacity and anatomy-level features can effectively guide the language decoder.

\subsection{Pathology-Report Alignment}

Existing report generation systems often lack interpretability and transparency, hindering their adoption in AI-aided diagnostics. Our model leverages the multi-label lesion detector to not only identify multiple pathologies within a single anatomical region but also serve as an intermediate output that offers a transparent mechanism for tracing and understanding the system's decisions, thereby enhancing reliability and credibility. This capability is crucial, especially when the model generates erroneous outputs or ambiguous findings. Should discrepancies arise, clinicians can pinpoint inaccuracies by reviewing the detected lesions, thus boosting the utility and trustworthiness of the system.

As depicted in Fig. \ref{fig:more-example}, the generated reports of our model highly align with the intermediate detection results. For instance, in the left example, the detector accurately identifies a lung lesion and mass/nodule, assigned to the right lung for prompt construction. Consequently, the generated report precisely notes a nodular opacity in the right lung, illustrating how our pathology-aware regional prompts improve the conveyance of anatomical and pathological information, thereby enhancing clinical accuracy.

Moreover, in the right example, a mild pulmonary edema is incorrectly reported by our model. This error can be traced back to the model mistakenly detecting this condition in multiple areas, which subsequently propagated the erroneous information. Overall, the consistency observed between the generated report and detection results underscores the potential of our lesion detector as an intermediary tool for tracing and explaining errors, thus enhancing trust and reliability in the technology.

\subsection{Limitations and Future Works}

Although our model performs satisfactorily, there is room for improvement. Currently, it only processes radiological image data, neglecting valuable textual diagnostic details. Incorporating image-text alignment techniques could enhance cross-modal integration and improve the quality of generated reports. Moreover, the accuracy of our model largely depends on the quality and diversity of the training data, with underrepresented pathologies or rare anatomical anomalies often leading to less accurate reports. Future improvements should focus on expanding the dataset to include a more balanced distribution of pathologies.

Additionally, the language decoder used in this study has a relatively small parameter count of around 2B, restricting the complexity of the prompt guidance and the model's generalization capabilities. The success of Vision-Language Foundation Models \cite{chexagent,medimageinsight,moor2023foundation} indicates that integrating LLMs as report decoders, alongside more expansive and diverse datasets, can significantly enhance model performance across various tasks. This strategy could yield more nuanced interpretations of medical images and lead to more precise diagnostic reports, thereby advancing the field of automated medical image analysis.

\section{Conclusion}

In this work, we propose a novel radiology report generation system that effectively harnesses both anatomical and pathological information through pathology-aware regional prompts, enhancing clinical accuracy. Unlike standard visual encoders that rely on patch-level features, our system includes an anatomical region detector to facilitate anatomy-level feature extraction under various scales, gaining holistic diagnostic views. Moreover, we introduce a multi-label lesion detector that identifies pathologies globally, significantly enhancing the reporting process. The prompt guidance is crafted by linking detected pathologies with overlapping anatomical regions, thus explicitly guiding the report decoder with a clinically relevant representation. Comprehensive quantitative and qualitative experimental results, along with formal expert evaluations, underscore the superiority of our model over previous state-of-the-art methods. The ablation studies further validate the effectiveness of our proposed prompt guidance.

\section*{References}
\vspace{-12pt}
\bibliographystyle{IEEEtran}
\bibliography{reference}

\end{document}